%% file: main.tex
\documentclass[10pt,twocolumn,letterpaper]{article}

\usepackage{cvpr}              

\input{preamble}

\definecolor{cvprblue}{rgb}{0.21,0.49,0.74}
\usepackage[pagebackref,breaklinks,colorlinks,allcolors=cvprblue]{hyperref}

\def\paperID{*****} 
\def\confName{CVPR}
\def\confYear{2026}

\title{Seeking Consensus: Geometric-Semantic On-the-Fly Recalibration for Open-Vocabulary Remote Sensing Semantic Segmentation}

\author{Guanchun Wang$^1$, Chenxiao Wu$^1$, Xiangrong Zhang$^1$$^{(\textrm{\Letter})}$, Zelin Peng$^2$, Jianxun Lai$^1$, \and Tianyang Zhang$^1$, Xu Tang$^1$\\
$^1$School of Artificial Intelligence, Xidian University\\
$^2$MoE Key Lab of Artificial Intelligence, AI Institute, Shanghai Jiao Tong University\\
\small{\tt{\{wangguanchun, xrzhang\}@mail.xidian.edu.cn}} \\ 
}

\begin{document}
\maketitle
\input{sec/0_abstract}    
\input{sec/1_Introduction}
\input{sec/2_Related_Work}
\input{sec/3_Proposed_Method}
\input{sec/4_Experiments_and_Analysis}
\input{sec/5_Conclusion}
{
    \small
    \bibliographystyle{ieeenat_fullname}
    \bibliography{main}
}


\end{document}

%% file: preamble.tex
\usepackage{soul}
\usepackage{url}
\usepackage{amsthm}
\usepackage{algorithm}
\usepackage{algorithmic}
\usepackage{marvosym}
\usepackage[most]{tcolorbox} 
\usepackage{multirow}
\usepackage{arydshln} 
\usepackage{colortbl} 

\definecolor{grayblue}{rgb}{0.184, 0.332, 0.589}
\definecolor{yellow}{rgb}{0.746, 0.563, 0.0}
\definecolor{green}{rgb}{0.402, 0.566, 0.25}

\newcommand{\gr}{\rowcolor[gray]{.95}}
\newcommand{\red}{\rowcolor{red!10}}
\newcommand{\blue}{\rowcolor{cyan!10}}
\newcommand{\yellow}{\rowcolor{yellow!10}}

\usepackage{xcolor}
\definecolor{darkpurple}{RGB}{74, 52, 90}
\definecolor{bluepurple}{RGB}{58, 144, 115}
\definecolor{darkbluegreen}{RGB}{81, 152, 146}
\definecolor{emerald}{RGB}{134, 171, 64}
\definecolor{olivegreen}{RGB}{93,162,100}
\definecolor{oliyellow}{RGB}{194,185,69}
\newcommand{\cbox}[1]{\textcolor{#1}{\rule{1.5ex}{1.5ex}}}

\definecolor{vaiGray}{RGB}{128, 128, 128}    
\definecolor{vaiRed}{RGB}{255, 0, 0}         
\definecolor{vaiLightGreen}{RGB}{144, 238, 144} 
\definecolor{vaiForestGreen}{RGB}{34, 139, 34}  
\definecolor{vaiBlue}{RGB}{0, 0, 255}        
\definecolor{vaiPurple}{RGB}{128, 0, 128}    

\definecolor{vddBlack}{RGB}{0, 0, 0}       
\definecolor{vddRed}{RGB}{255, 0, 0}       
\definecolor{vddYellow}{RGB}{255, 255, 0}  
\definecolor{vddGreen}{RGB}{0, 255, 0}     
\definecolor{vddBlue}{RGB}{0, 0, 255}      
\definecolor{vddMagenta}{RGB}{255, 0, 255} 
\definecolor{vddCyan}{RGB}{0, 255, 255}    
\definecolor{uddGreen}{RGB}{34, 139, 34}   
\definecolor{uddRed}{RGB}{255, 0, 0}       
\definecolor{uddGray}{RGB}{128, 128, 128}  
\definecolor{uddBlue}{RGB}{0, 0, 255}      
\definecolor{uddBlack}{RGB}{0, 0, 0}       

\tcbset{
    findingbox/.style={
        colback=black!10,
        colframe=black!60,
        fonttitle=\bfseries,
        left=2pt,
        right=2pt,
        top=2pt,
        bottom=2pt,
        arc=4pt,
        outer arc=4pt,
        boxrule=0.5pt,
        before upper={\parindent=0pt},
    }
}



%% file: sec/0_abstract.tex
\begin{abstract}
  Open-vocabulary semantic segmentation (OVSS) in remote sensing images is a promising task that employs textual descriptions for identifying undefined land cover categories. Despite notable advances, existing methods typically employ a static inference paradigm, overlooking the distinct distribution of each scene, resulting in semantic ambiguity in diverse land covers and incomplete foreground activation. Motivated by this, we propose \underline{\textbf{See}}king \underline{\textbf{Co}}nsensus, termed \textbf{SeeCo}, a plug-and-play framework to boost the performance of training-free OVSS models in remote sensing images, which recalibrates arbitrary OVSS models on-the-fly by seeking dual consensus: geometric consensus learning (GCL) through multi-view consistent observations and semantic consensus learning (SCL) via textual description adaptive calibration, which assists collaborative recalibration of visual and textual semantics. The two consensus are injected via an online consensus injector (OCI), effectively alleviating the under-activation and semantic bias. SeeCo requires no specific training process, yet recalibrates semantic-geometric alignment for each unique scene during inference. Extensive experiments on eight remote sensing OVSS benchmarks show consistent gains, proving its effectiveness and universality. 
\end{abstract}

%% file: sec/1_Introduction.tex
\section{Introduction}


\label{sec:intro}
Remote sensing semantic segmentation, a foundational task in Earth observation, plays an essential role in practical applications such as precision agriculture, hazard assessment, and urban planning \cite{ijcaiss2,ACMMM-SS,ACMMM-RS1}. With the progress of vision language models (VLMs), open-vocabulary semantic segmentation (OVSS) \cite{TACOSS,HyperSP} has attracted growing research awareness. By leveraging textual descriptions and pretrained VLMs, OVSS takes a detour from the traditional paradigms that rely on a closed set of pre-defined categories, providing a feasible solution for zero-shot land cover extraction from remote sensing images.
\begin{figure}[t]
  \centering
  \includegraphics[width=\linewidth]{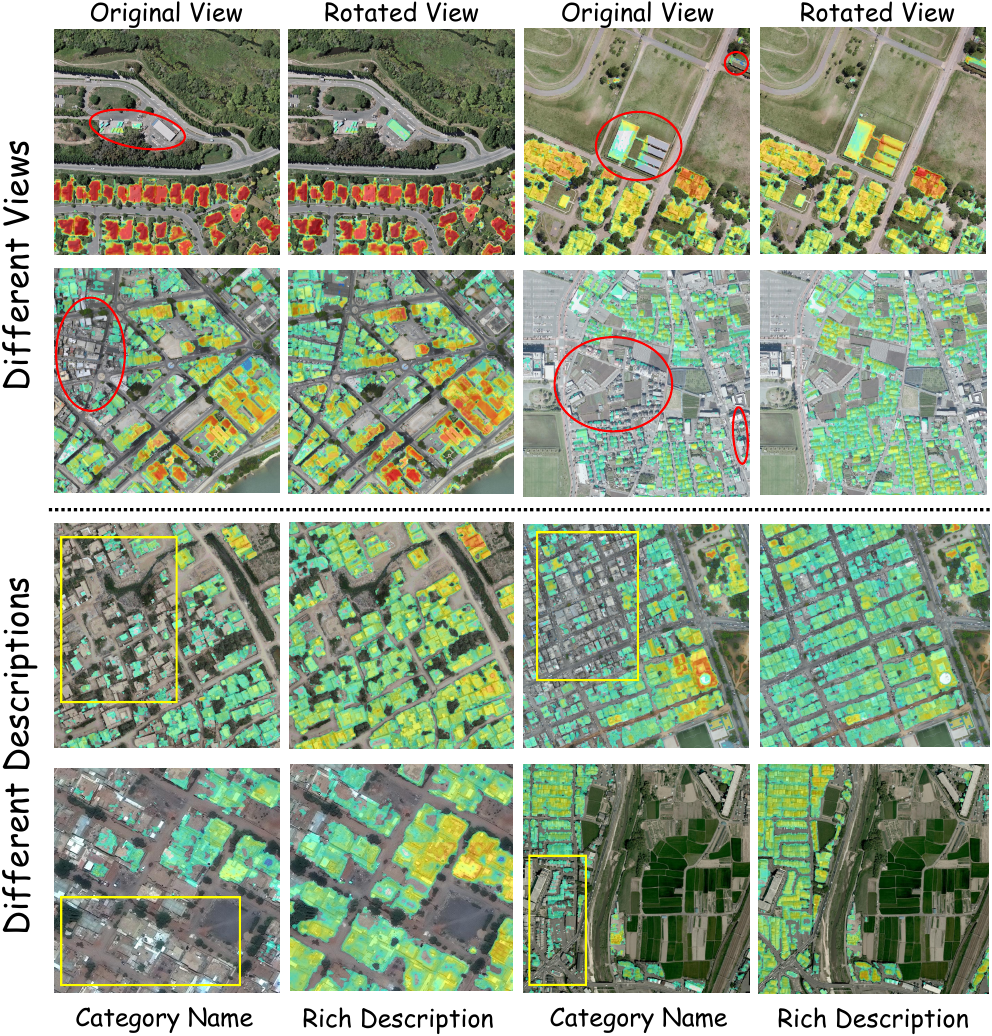}
  \caption{Comparison of open-vocabulary semantic segmentation results in remote sensing scenes from existing SOTA methods \cite{ProxyCLIP}, under different observed views (top) and different text descriptions (bottom). The arbitrary orientation and intra-class heterogeneity render it challenging to fully capture targets via only a single view or a category name.}
  \label{intro1}
\end{figure}
Existing OVSS methods can be broadly divided into fine-tuning and training-free methods. The former \cite{CAT-Seg,HyperSP} first fine-tunes image-level VLMs, especially for CLIP \cite{CLIP}, on closed-set semantic segmentation datasets with dense annotations to enable segmentation capability, and then infers the open-domain images. The latter aims to introduce the ability of dense prediction via a series of training-free strategies, such as modifying the architecture of CLIP \cite{ClearCLIP}, or integrating CLIP with vision foundation models \cite{ProxyCLIP,SegEarth-OV,SegEarth-OV3}, which are more appropriate for remote sensing images that require high costs of pixel-level annotations. Despite promising progress via representation resolution enhancement, the above methods follow a static inference paradigm that ignores the unique distribution of each RSI, thereby leading to the following challenges in complex scenes:

Unlike natural scenarios, which are typically object-centric and have relatively fixed spatial arrangements, RSIs are captured from a bird’s eye view, resulting in land covers with arbitrary orientations and complex layouts, such as roads, buildings, and vehicles. As illustrated in Figure \ref{intro1}, the static inference paradigm, constrained to a single observation perspective, fails to accommodate geometric variations, leading to incomplete activation in object regions. Therefore, we argue the following design principle: 

\begin{tcolorbox}[findingbox]
\textbf{Guideline 1:} A remote sensing OVSS framework should maintain rotation-invariant representation capabilities to match observations from bird's-eye views.
\end{tcolorbox}

Land covers exhibit substantial intra-class heterogeneity. For instance, the category “building” encompasses both densely packed residential structures and sparse skyscrapers, resulting in a remarkable difference in visual appearance. Static textual descriptions further harm cross-modal matching in vision-language models, which are mainly pretrained on natural scenes while dealing with remote sensing captions, leading to semantic bias, as shown in Figure \ref{intro1}. Previous studies \cite{HyperSP} state that maintaining the text encoder of CLIP frozen can enhance generalization on open-vocabulary tasks, thereby retaining its robust textual embeddings across numerous categories. Hence, we formulate another design principle as follows:

\begin{tcolorbox}[findingbox]
\textbf{Guideline 2:} A remote sensing OVSS framework should alleviate its semantic bias while keeping the text encoder frozen.
\end{tcolorbox}

Motivated by the above analysis, we propose \textbf{\textit{Seeking Consensus}}, termed \textbf{\textit{SeeCo}}, a plug-and-play and inference-time enhancement framework that can ameliorate most existing training-free OVSS methods. By simultaneously aligning semantic and geometric consensus during inference, SeeCo takes a detour from the limitations of static inference paradigms and realizes the dynamic adaptation of visual features and textual semantics, thereby improving remote sensing OVSS performance. 

More specifically, inspired by the multi-view observation of remote sensing images, we first introduce a geometric consensus learning (GCL) module that enforces rotation-variant self-supervised regularization, facilitating the model to produce consistent object representations across multi-orientation views, which significantly enhances the geometric robustness of object region features and alleviates the under-activation issue. Next, considering that a single category text cannot adequately capture land covers with intra-class diversity, we propose a semantic consensus learning (SCL) mechanism that mines rich knowledge through a multi-modal collaborative prompting strategy and dynamically recalibrates textual descriptions based on the distribution of each scene, thereby reducing semantic bias without tuning the frozen text encoder. Finally, we design a lightweight online consensus injector (OCI) that integrates the above consensus knowledge into the model during inference, enabling on-the-fly adaptive enhancement.

The main contributions of this work can be summarized as follows:

\begin{itemize}
	\item We propose SeeCo, a plug-and-play framework for remote sensing OVSS through seeking geometric and semantic consensus on-the-fly during inference, which leads to substantial improvements in segmentation results by leveraging scene-specific feature distributions.
    
	\item We develop a geometric consensus learning mechanism (GCL) and a semantic consensus learning mechanism (SCL) via rotation consistency-based regularization and textual description adaptive calibration, which assist collaborative recalibration of visual and textual semantics. The two consensus are injected online via a lightweight online consensus injector (OCI), effectively mitigating the issues of under-activation and semantic bias.
    
	\item Extensive experiments on eight public remote sensing semantic segmentation benchmarks demonstrate that SeeCo, without any offline training or manual annotation, significantly outperforms state-of-the-art methods. 
\end{itemize}

%% file: sec/2_Related_Work.tex
\section{Related Work}
\subsection{Open-Vocabulary Semantic Segmentation}
Existing methods on open-vocabulary semantic segmentation are broadly categorized into fine-tuning and training-free paradigms. 

\paragraph{{Fine-tuning OVSS Methods.}} The fine-tuning OVSS paradigm follows the manners of mask proposal classification \cite{OpenSeg,ZegFormer} or dense feature adaptation \cite{LSeg,SED,CAT-Seg} to transfer the cross-modal representation in vision-language models (e.g., CLIP \cite{CLIP}) to segmentation tasks. The former \cite{OpenSeg,ZegFormer} first organizes pixels into class-agnostic groups and then aligns regional vision features with text embeddings for OVSS. The latter \cite{LSeg,SED,CAT-Seg} aims to directly perform pixel-level alignment between visual and textual features from vision-language models and enhances the semantic consistency by introducing multi-scale feature extraction or global context correspondence maps. Considering the properties of RSIs, previous studies improve OVSS through feature enhancement \cite{GSNet}, domain-specific pre-training \cite{SkySense-O}, and foundation model integration \cite{AerOSeg,SCORE}. Although such paradigms exhibit remarkable results, they rely on numerous pixel-level annotations and high computational costs, restricting rapid deployment in remote sensing applications.

\paragraph{Training-free OVSS Methods.} 
Training-free OVSS paradigms have gained increasing attention, aiming to directly exploit the inherent discriminative representation capacity of pre-trained VLMs without additional training, which fall into two technical streams. One stream focuses on refining the intrinsic feature representations of CLIP via spatial coherence of attention maps \cite{SC-CLIP} or pruning redundant layers \cite{ClearCLIP}. Another stream incorporates vision foundation models \cite{DINO,SAM} to provide spatial priors \cite{ProxyCLIP,CorrCLIP} for boundary refinement. Similarly, existing OVSS methods for remote sensing images mainly focus on the training-free paradigm and also attempt to improve segmentation results via spatial feature enhancement or integrating vision foundation models. SegEarth-OV \cite{SegEarth-OV} builds a self-supervised feature upsampling module to recover fine details of segmentation maps, realizing outstanding performance. InstructSAM \cite{InstructSAM} interprets user instructions and generates mask proposals via a large language model and VFM \cite{SAM}, respectively, which formulates label assignment as an optimization problem, achieving promising results. Building upon the more recent VFM \cite{SAM3}, SegEarth-OV3 \cite{SegEarth-OV3} directly fuses outputs from its semantic and instance heads, filtering redundant categories via presence scores.

\textit{Despite promising progress, these methods still follow static inference paradigms, which struggle to dynamically adapt to scene-specific images under varying observations and complex intra-class heterogeneity}.
\begin{figure*}[!t]
  \centering 
  \includegraphics[width=\linewidth]{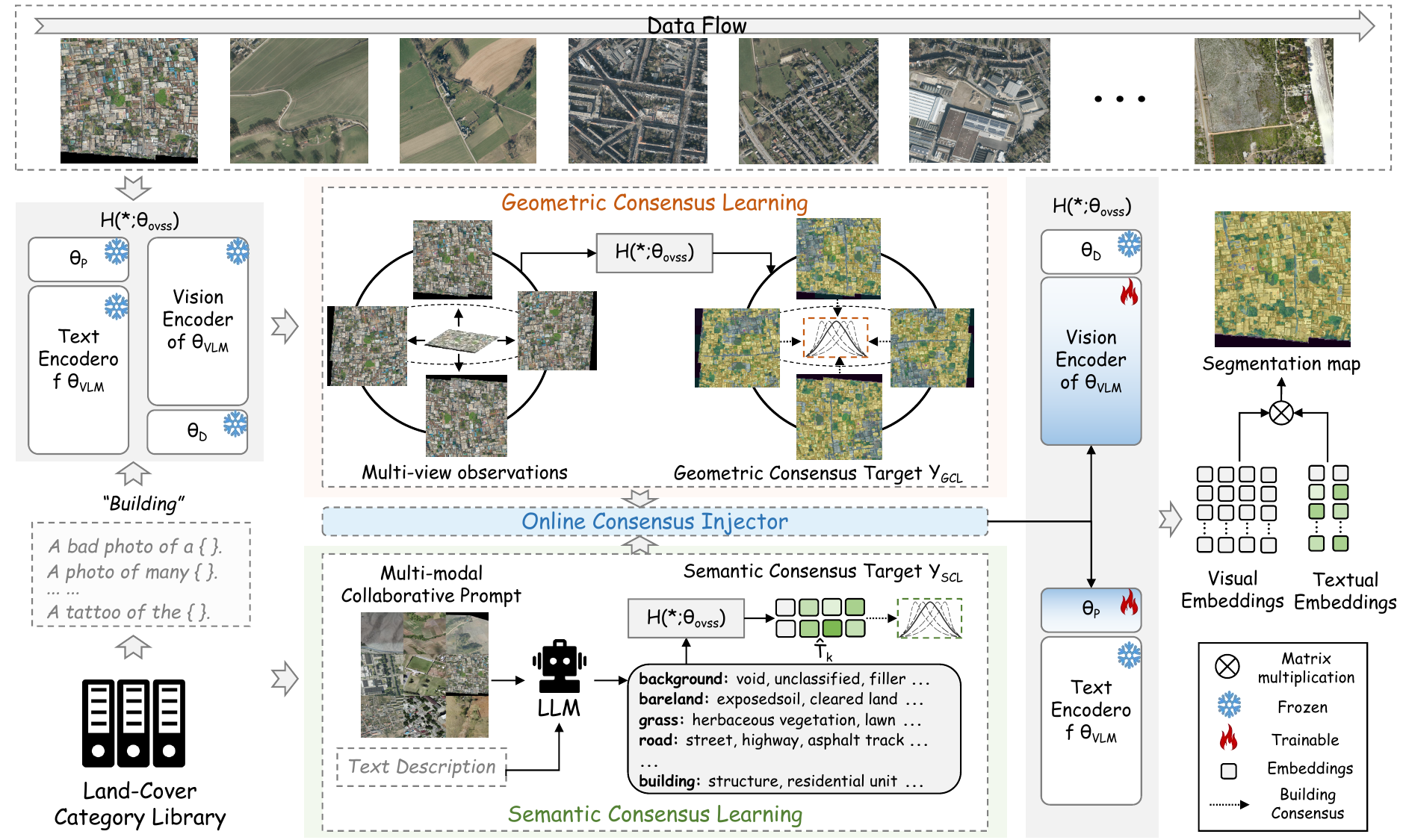}
  \caption{Illustration of SeeCo that first leverages a geometric consensus learning (GCL) module to simulate different observation views and acquire multi-view prediction maps for building geometric consensus, then adopts a semantic consensus learning (SCL) module for adaptively enriching textual descriptions via multi-model collaborative prompting, yielding semantic consensus. Next, an online consensus injector (OCI) is used to integrate the above consensus for scene-specific model adaptation, enhancing segmentation performance.}
  \label{Overall Framework}
\end{figure*}

\subsection{Test-Time Training}
Our proposed method is related to test-time training (TTT), which dynamically updates the model during inference to adapt to distribution shifts between training and test data. The core challenge of TTT lies in designing effective optimization objectives for unlabeled test samples while ensuring update efficiency. Previous research follows two primary technical solutions: Some previous studies \cite{TENT} focus on statistical adjustment of normalization layers using entropy minimization, thereby narrowing the gap between training and testing distributions. Other works construct self-supervised auxiliary tasks \cite{Sun-etal,MaskSSL}, where proxy objectives like masked reconstruction are introduced during inference to facilitate the adaptation of test data. Moreover, with the advance of VLMs, a series of test-time prompt tuning \cite{TPT,C-TPT} or reinforcement learning-based approaches \cite{RLCF} have emerged, which enhance cross-domain robustness by applying diverse data augmentations to individual test samples and minimizing marginal prediction entropy. In this work, we take inspiration from TTT to break the static inference paradigm of existing OVSS methods and guide the model to learn scene-specific attributes.

%% file: sec/3_Proposed_Method.tex
\section{Proposed Method}
\label{sec:method}

\subsection{Preliminaries}
\label{subsec:preliminaries}

\subsubsection{{Training-free Open-Vocabulary Semantic Segmentation}}

Training free open-vocabulary semantic segmentation (TF-OVSS) aims to directly perform pixel-level classification of undefined categories without retraining VLM with dense annotations. Specifically, given an input image $\mathbf{I}\in \mathbb{R}^{H\times W\times C}$ and a set of open-vocabulary categories $\mathcal{C} = \{c_1, c_2, \dots, c_J\}$, a frozen VLM is first utilized to extract dense features of images and text embeddings of each category description:
\begin{equation}
\begin{aligned}
    \mathbf{V} &= \mathcal{H}(\mathcal{H}(\mathbf I; \Theta_\text{VLM});\Theta_{\text{D}})\in  \mathbb{R}^{H\times W\times D}\\
    \mathbf{T} &=  \mathcal{H}(\mathcal H(c_j; \Theta_{\text{P}});  \Theta_\text{VLM})\in  \mathbb{R}^{D\times J}
\end{aligned},
\end{equation}
where $\Theta_{\text{VLM}}$ and $\Theta_{\text{D}}$ indicate the frozen VLM and dense feature mapping module. $\mathcal H(*; \Theta_{\text{P}})$ is a prompt mechanism for each category. Then, a cross-modal matching operation $\mathcal{S}$ is adopted to calculate the similarity between the above features, thereby acquiring the predicted label for each spatial location.
\begin{equation}
    \hat{\mathbf Y}_{{ p}} =\arg\max_{j} \, \mathcal{S}( \mathbf{V}_{{p}}, \mathbf{T}_j ),
\end{equation}
where $\hat{\mathbf Y}_{{p}}$ denotes the predicted label of the pixel $p$. Existing methods majorly focus on designing more powerful mapping functions $\mathcal{H}(*;\Theta_{\text{D}})$ by integrating VFMs \cite{ProxyCLIP} or feature upsampling modules \cite{SegEarth-OV} to enhance dense representation. However, they typically follow a static inference paradigm, where $\mathcal{H}(*;\Theta_{\text{D}})$ and $\mathcal H(*; \Theta_{\text{P}})$ are fixed across different test images, which overlooks the unique property of each scene and the gap between VLM and land cover categories in RSIs, impairing cross-modal recognition. To alleviate it, we dynamically optimize the above terms for each image during inference.

\subsubsection{Test-Time Training}
Test-time training aims to improve the generalization capability of models under unseen data distributions, whose core process is to adjust model parameters during inference by self-supervised auxiliary tasks, mitigating the distribution shift between training and testing data. For each unlabeled test sample $\mathbf{X}$, the model is updated by minimizing the self-supervised loss constructed by $\mathbf{X}$ itself:
\begin{equation}
    \Theta_{\mathbf X} = \arg\min_{\Theta} \, \mathcal{L}_{\text{ssl}}^\mathbf{X}( \mathbf{X}; \Theta ),
\end{equation}
where $\Theta_{\mathbf X}$ denotes the updated parameters specifically to the test sample $\mathbf{X}$. Then, these sample-specific parameters are leveraged for inferring the main task.

\subsection{Seeking Consensus}
\subsubsection{\textbf{Overall Pipeline}}
For an arbitrary training-free OVSS model $\mathcal H(*;\Theta_\text{VLM},\Theta_\text{D},\Theta_{\text{P}})$ illustrated in Figure \ref{Overall Framework}, given a test image $\mathbf I$ and a land-cover category name $c_j$, SeeCo first feeds $I$ into a geometric consensus learning module, which simulates multi-view observation via rotational geometric transformations $\mathcal G(*)$ and yields a series of segmentation maps $\mathcal{Y}={\{ \hat{\mathbf{Y}}^k\}_{k=1}^{K}}$. A robust geometric consensus target $\mathbf Y_{\text{GCL}}$ is then constructed by aggregating these multi-view predictions. Next, the category name $c_j$ is fed into a semantic consensus learning module, which leverages a multi-modal collaborative prompting strategy for large language models to generate multiple descriptions for each land cover category to enrich textual representation. Both the original $c_j$ and enriched descriptions $\bar c_j$ are encoded using $\mathcal H(\mathcal H(*;\Theta_{\text{P}});\Theta_\text{VLM})$, resulting in a robust semantic consensus target $\mathbf Y_{\text{SCL}}$.

Subsequently, a online consensus injector is established, which builds a lightweight parameter tuning branch and an adaptive prompt fusion module for the vision and text encoders, respectively. OCI is optimized online in a self-supervised manner using geometric consensus $\mathbf Y_{\text{GCL}}$ and semantic consensus $\mathbf Y_{\text{SCL}}$ as supervision signals, whose loss function is defined as:
\begin{equation}
\label{eq4}
    \mathcal{L}_{\text{SeeCo}} = \frac{1}{K} \sum_{k=1}^{K} {\Phi}(\mathbf Y_{\text{GCL}}, \hat{\mathbf Y}^{k}) + \sum_{\breve{\mathbf Y}\in\{\hat{\mathbf Y},\bar {\mathbf Y}\}} \Phi(\mathbf Y_{\text{SCL}},\breve{\mathbf Y}),
\end{equation}
where $\Phi$ is mean square error. After test-time training, we adopt $\Theta_{\text{SeeCo}}$ to calculate the final segmentation map by integrating geometric and semantic consensus predictions:
\begin{equation}
    \hat {\mathbf Y} = \arg\max_{k \in \mathcal{C}} \left( \delta \cdot \mathbf Y_{\text{GCL}} + (1-\delta) \cdot \mathbf Y_{\text{SCL}} \right),
\end{equation}
where $\delta \in [0,1]$ is the trade-off to balance the contributions between geometric and semantic consensus.

\subsubsection{\textbf{Geometric Consensus Learning}}
Considering the land cover in remote sensing images exhibits arbitrary orientations due to its bird's-eye view, existing static models often produce inconsistent activations under different observations. To mitigate this geometric ambiguity, we design a geometric consensus learning module to learn the rotation invariance of geospatial objects via a self-supervised auxiliary task constraint. 

Specifically, given an image $\mathbf{I}$, we apply a set of rotation operators $\mathcal{G}(*;\theta_{\text{obs}})$ to obtain $K$ observations as follows:
\begin{equation}
    {\mathcal{I}}=\{\mathcal G(\mathbf I,\theta_{\text{obs}}^k)|\theta_{\text{obs}}^k=\frac{2k\pi}{K},k\in[1,K] \}
\end{equation}
where $\mathcal I$ denotes a set of multi-view images. Each observed view is processed by the frozen segmentation model to yield a series of segmentation probability maps:
\begin{equation}
    {\mathcal{Y}}=\{
    \mathcal{S}( \mathcal H(\mathcal H(\mathbf{I}_\text{GCL}; \Theta_\text{VLM});\Theta_\text{D}), \mathbf{T} )|\mathbf{I}_\text{GCL}\in\mathcal{I}\}
\end{equation}

Next, we perform the inverse transformation operation $\mathcal{G}^{-1}(*;\theta_{\text{obs}})$ of such observations to calculate the geometric consensus target:
\begin{equation}
    {\mathbf{Y}}_{\text{GCL}} = \frac{1}{K} \sum_{k=1}^{K} \mathcal{G}^{-1}(\mathcal{Y}^k;\theta_{\text{obs}}^k)
\end{equation}
where the geometric consensus target $\hat{\mathbf{Y}}_{\text{GCL}}$ serves as a robust pseudo-mask to recalibrate the model online.

\subsubsection{\textbf{Semantic Consensus Learning}}
Due to the high intra-class heterogeneity in remote sensing scenes, existing methods using VLMs trained on natural scenes often fail to capture diverse visual semantic expressions, while finetuning the text encoder of VLMs will reduce its generalization ability. We thus propose a semantic consensus learning module to adaptively recalibrate the textual embedding without tuning the text encoder, improving the foreground activation of diverse land covers.

\begin{figure}[t]
    \centering
    \resizebox{\linewidth}{!}{
    \begin{tcolorbox}[findingbox]
    \begin{minipage}[t]{0.57\textwidth}
        \begin{center}
        \textbf{Textual knowledge prompt:}
        \end{center}
        I will use the vision-language model for semantic segmentation of remote sensing images. Referring to the attached scene, please provide synonyms for the given category in the current scene. For example:"large vehicle: truck, lorry, bus, heavy vehicle, transport vehicle"
        \end{minipage}
        \hfill
        \begin{minipage}[t]{0.42\textwidth}
        \begin{center}
        \textbf{Visual prior prompt:}
        \end{center}
        \includegraphics[width=\textwidth]{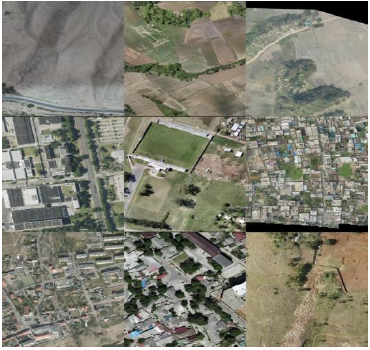}
        \end{minipage}
    \end{tcolorbox}}
    \caption{Illustration of multi-modal collaborative prompting, which adopts textual knowledge and visual prior prompts to generate more category descriptions.}
  \label{prompt}
\end{figure}

{\textbf{Multi-modal Collaborative Prompting.}} SCL first constructs a multi-modal collaborative prompting strategy for guiding LLMs \cite{GPT4} that contain rich external knowledge to enrich the textual descriptions for land cover categories. Inspired by \cite{Align}, we employ a multimodal question-answering manner to guide LLMs to produce diversified yet proper descriptions of remote sensing land cover categories, which consists of two parts: visual prior prompting and textual knowledge acquisition. The former randomly selects $L$ images $\{\mathbf I^l\}_{l=1}^L$ from the known dataset and feeds them into LLMs as visual scene priors, enabling LLMs to understand the scene where the land cover category exists. The latter conducts a conversation with LLMs using the statement $\mathbf Q$ to obtain the enriched library $\hat{\mathcal C}=\mathcal H(\mathbf Q|\{\mathbf I^l\}_{l=1}^L;\Theta_{\text{LLM}})$ consisting of $Z$ synonyms for each category, as illustrated in Figure \ref{prompt}.   

{\textbf{Adaptive Context Recalibration.}} Then, an adaptive context recalibration mechanism is built in SCL, which estimates a group of scene-adaptive contexts $\mathcal W(\mathbf I)\in\mathbb{R}^{D\times Z}$ to integrate enriched textual descriptions. For the $j$-th category, we extract its enriched textual descriptions ${\hat{\mathcal C}}_k$ and original descriptions $\mathcal C_k$ and feed them into the frozen text encoder of VLMs to form a series of textual embeddings $\{\mathbf T^j\in\mathbb{R}^{D\times 1}, \hat{\mathbf T}^j\in\mathbb{R}^{D\times Z}\}$. Next, scene-adaptive contexts $\mathcal W(\mathbf I)$ are leveraged to fuse these embeddings:
\begin{equation}
    \breve{\mathbf{T}}_j = \sum_{z=1}^{Z} \frac{\exp\left(\mathcal{W}(\mathbf{I})^z / \tau\right)}{\sum_{o=1}^{Z} \exp\left(\mathcal{W}(\mathbf{I})^o / \tau\right)} \cdot \hat{\mathbf{T}}_{j}^z,
\end{equation}
where $\tau$ is the temperature factor. The embeddings of enriched textual descriptions $\breve{\mathbf{T}}_j$ and original descriptions $\mathbf{T}_j$ are utilized to build a semantic consensus target $\mathbf Y_{\text{SCL}}=\mathcal{S}(\mathbf V, \mathbf{T} )+\mathcal{S}(\mathbf V, \breve{\mathbf{T}} )$ for model online adaptation.

\subsubsection{\textbf{Online Consensus Injector}}
To transfer the above consensus into existing OVSS models, we build a online consensus injector that is composed of two modules: scene-adaptive contexts and a lightweight parameter tuning branch to optimize text and vision encoders of OVSS models online, respectively, thereby learning semantic and geometric consensus knowledge. The former is claimed in the above section, which is a set of trainable embeddings associated with each category description, injecting the scene-specific knowledge into the text encoder.

The latter is a low-rank layer \cite{LoRA} that rebuilds the fully connected layers of the last $P$ Transformer blocks in the vision encoder. For the target layer $\mathbf{\Theta}_{\text{FC}}$, it can be rebuilt as $\mathcal {H}(\mathbf X;\mathbf{\Theta}_{\text{FC}}+\beta\cdot \Theta_\mathbf{B} \Theta_\mathbf{A})$. $\Theta_\mathbf{A}\in \mathbb{R}^{r \times D}$, $\Theta_\mathbf{B}\in \mathbb{R}^{D \times r}$ are low-rank trainable matrices with a rank of $r$ and a scaling factor of  $\beta$. Through the above designs, the consensus can be introduced into the original models during the inference.

%% file: sec/4_Experiments_and_Analysis.tex
\section{Experiments and Analysis}
\subsection{Datasets and Evaluation Metrics}
We evaluate our SeeCo on eight challenging remote sensing semantic segmentation datasets, including satellite and UAV imagery: OpenEarthMap \cite{OpenEarthMap}, LoveDA \cite{LoveDA}, iSAID \cite{iSAID}, Potsdam \cite{ISPRS}, Vaihingen \cite{ISPRS}, UAVid \cite{UAVid}, UDD \cite{UDD5}, and VDD \cite{VDD}. For a fair comparison, we keep the same category definitions as \cite{SegEarth-OV}. Detailed dataset descriptions and category definitions are provided in the supplementary material. We employ the mean Intersection over Union (mIoU) metric to evaluate the segmentation performance.
\begin{figure}[t]
    \centering
    \includegraphics[width=.98\linewidth]{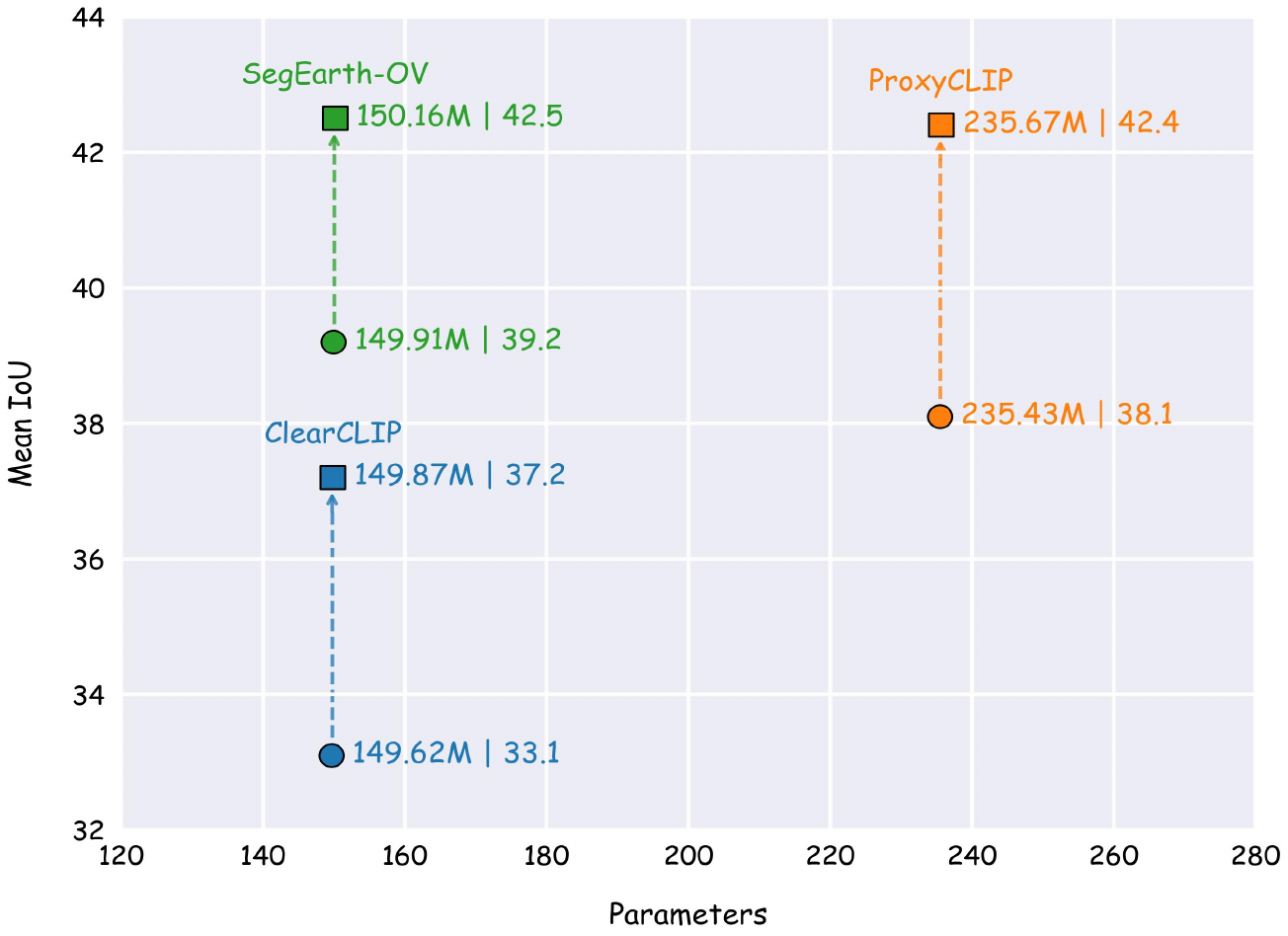}
    \caption{Comparisons of performance gains and parameters inserting SeeCo into existing methods on eight remote sensing open-vocabulary semantic segmentation datasets.}
    \label{gain}
\end{figure}
\begin{table*}[t]
    \centering
    \caption{Comparison with state-of-the-art methods on eight remote sensing open-vocabulary semantic segmentation datasets. The best results are marked in bold, and the suboptimal results are marked using an underline.} 
    \label{table_overall}
    \resizebox{\linewidth}{!}{
    \begin{tabular}{lccccccccc}
            \toprule
            Method & OpenEarthMap & LoveDA & iSAID & Potsdam & Vaihingen & UAVid$^{\text{img}}$ & UDD5 & VDD & Avg \\
            \midrule 
            \midrule 
            \gr \multicolumn{10}{c}{ \textbf{\textit{Training-free OVSS Methods}} }\\
            CLIP$_{\text{ICML2021}}$ \cite{CLIP} & 12.0 & 12.4 & 7.5 & 14.5 & 10.3 & 10.9 & 9.5 & 14.2 & 11.4 \\
            MaskCLIP$_{\text{ECCV2022}}$ \cite{MaskCLIP} & 25.1 & 27.8 & 14.5 & 31.7 & 24.7 & 28.6 & 32.4 & 32.9 & 27.2 \\
            GEM$_{\text{CVPR2024}}$ \cite{GEM} & 33.9 & 31.6 & 17.7 & 36.5 & 24.7 & 33.4 & 41.2 & 39.5 & 32.3 \\
            SCLIP$_{\text{TIP2025}}$ \cite{SC-CLIP} & 29.3 & 30.4 & 16.1 & 36.6 & 28.4 & 31.4 & 38.7 & 37.9 & 31.1 \\
            CorrCLIP$_{\text{ICCV2025}}$ \cite{CorrCLIP} & 32.9 & 36.9 & \textbf{25.5} & \textbf{51.9} & \textbf{47.0} & 38.3 & 46.1 & \underline{47.3} & 40.7 \\
            ClearCLIP$_{\text{ECCV2024}}$ \cite{ClearCLIP} & 30.8 & 31.0 & 18.2 & 40.4 & 27.3 & 36.7 & 41.8 & 38.8 & 33.1 \\
            ProxyCLIP$_{\text{ECCV2024}}$ \cite{ProxyCLIP} & 39.2 & 36.4 & 21.4 & 44.9 & 30.6 & 40.9 & 47.5 & 43.9 & 38.1 \\
            SegEarth-OV$_{\text{CVPR2025}}$ \cite{SegEarth-OV} & 40.3 & 36.9 & 21.7 & 47.1 & 29.1 & 42.5 & 50.6 & 45.3 & 39.2 \\
            \midrule
            \gr \multicolumn{10}{c}{ \textbf{\textit{Our Proposed Method}}} \\
            \red SeeCo+ClearCLIP$_{\text{ECCV2024}}$ & 34.8 & 36.6 & 19.5 & 43.5 & 37.5 & 40.0 & 44.5 & 40.9 & 37.2 \\
            \blue SeeCo+ProxyCLIP$_{\text{ECCV2024}}$ & \underline{42.9} & \textbf{40.3} & \textbf{22.7} & \underline{49.8} & \textbf{42.5} & \underline{44.3} & \underline{50.3} & \textbf{46.7} & \underline{42.4} \\
            \yellow SeeCo+SegEarth-OV$_{\text{CVPR2025}}$ & \textbf{43.0} & \underline{39.4} & \underline{22.2} & \textbf{50.4} & \underline{41.2} & \textbf{44.9} & \textbf{53.6} & \underline{45.3} & \textbf{42.5} \\
            \bottomrule
    \end{tabular}
    }
\end{table*}
\subsection{Implementation Details}
All experiments are implemented on a single NVIDIA RTX 4090 GPU. We follow the original image preprocessing settings of each baseline to preserve their segmentation performance. All methods employ the pre-trained CLIP (ViT-B/16) as the multi-modal feature extractor $\Theta_{\text{VLM}}$. During the testing phase, SeeCo first serves Eq.~\ref{eq4} as the objective function to learn scene-specific consensus knowledge, and then employs the AdamW optimizer with a learning rate of $3 \times 10^{-4}$ to update OCI for a single iteration. During inference, all models adopt the same sliding-window strategy with a window size of $224 \times 224$ and a stride of $112$. In geometric consensus learning, the number of multi-view observations $K$ is set to 4. In semantic consensus learning, scene-adaptive contexts $\mathcal W$ are initialized to zero, and the temperature factor $\tau$ is set to 0.01. In online consensus injector, $P$ is set to 2, the rank of two low-rank matrices is 8, and the scaling factor is set to 16. The consensus factor $\delta$ is set to 0.5.

\subsection{Comparison with State-of-the-arts}
Since SeeCo is a plug-and-play framework, we select two general OVSS methods \cite{ClearCLIP,ProxyCLIP} and a remote sensing OVSS method \cite{SegEarth-OV} as the basic models. 
\begin{figure}[t]
    \centering
    \includegraphics[width=.9\linewidth]{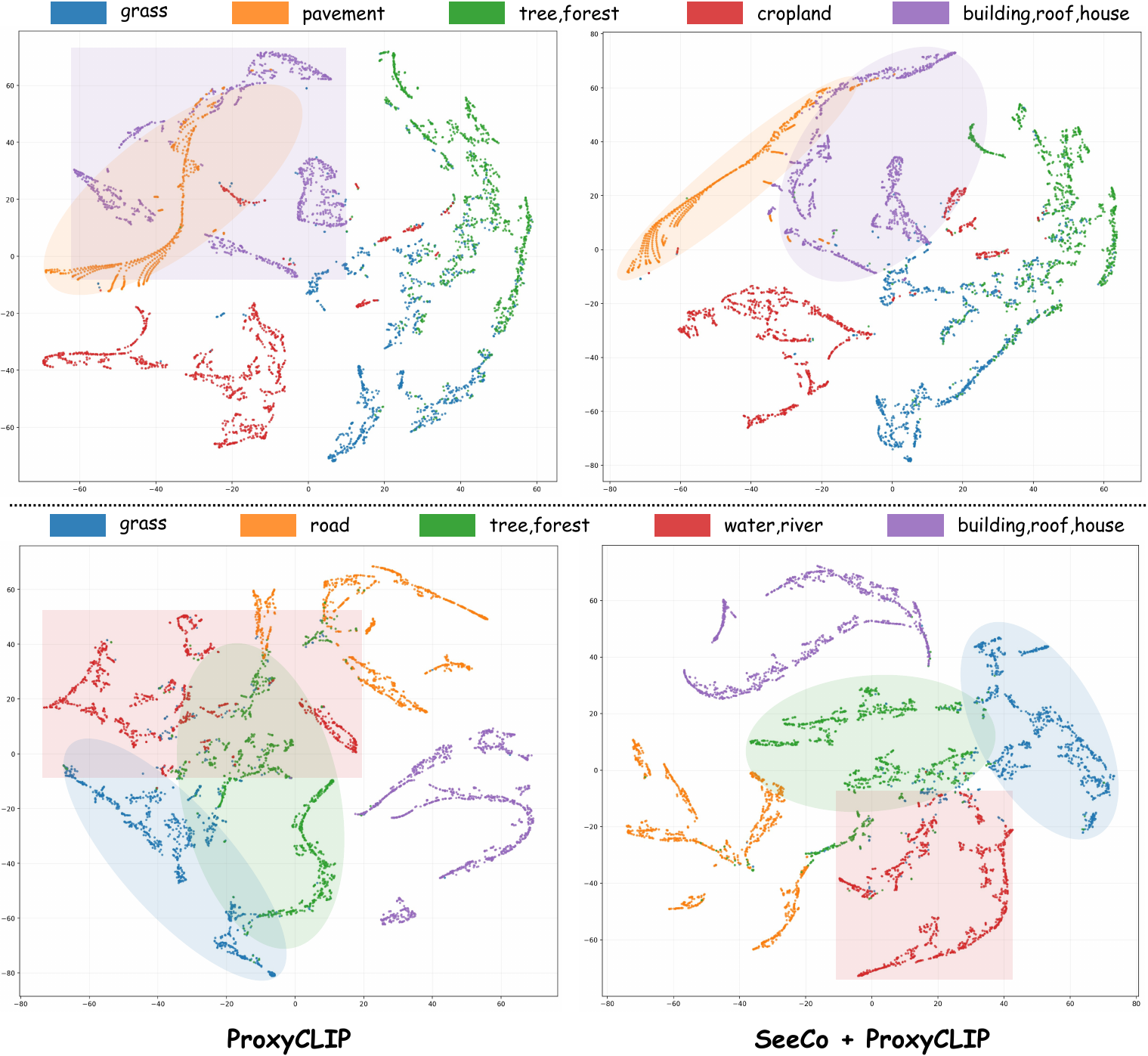}
    \caption{t-SNE feature distribution comparison on OpenEarthMap with and without SeeCo.}
    \label{tsne}
\end{figure}

\begin{figure}[t]
    \centering
    \includegraphics[width=\linewidth]{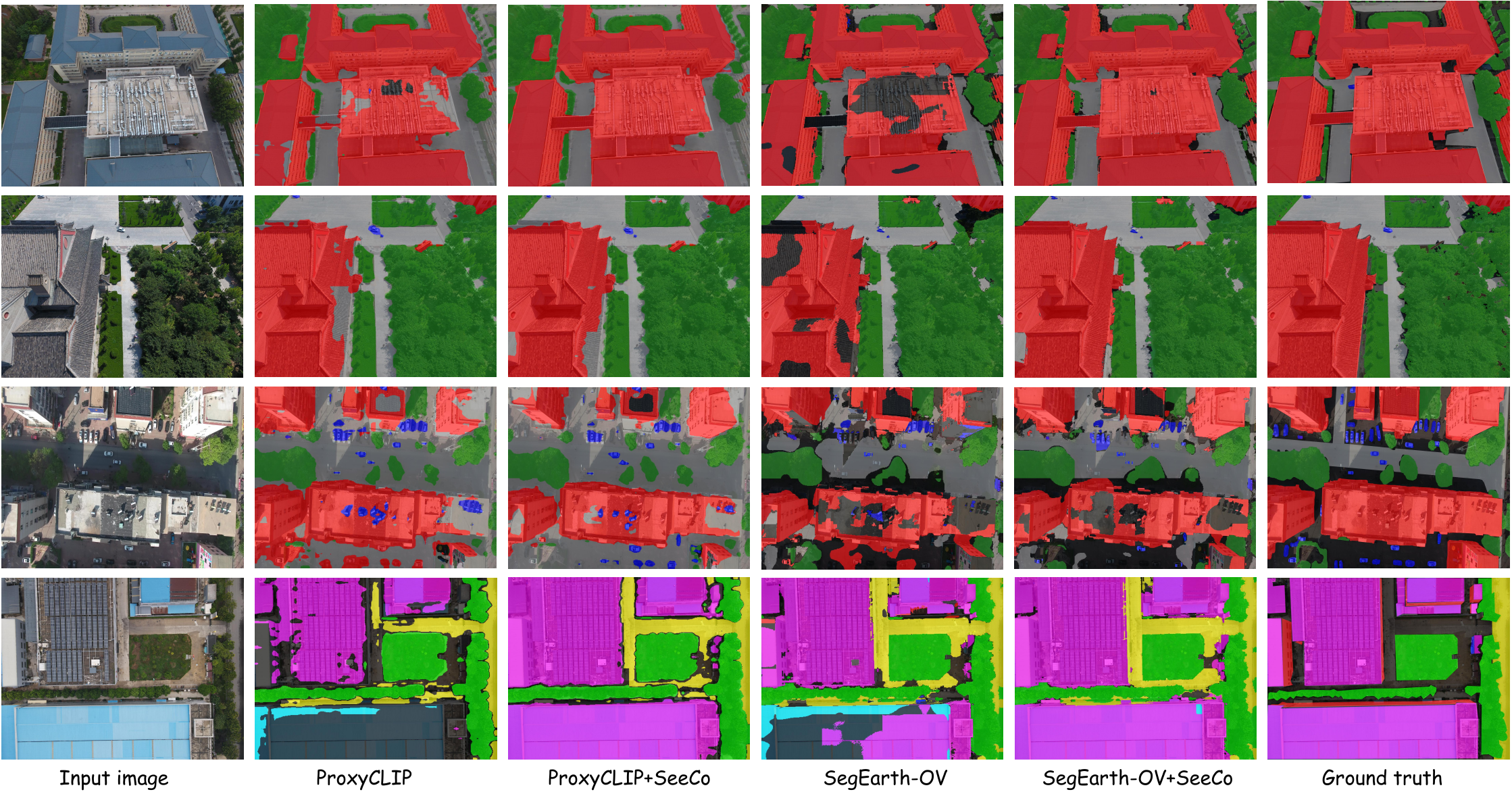}
    \caption{Qualitative results on UDD5 and VDD. 
        UDD5: 
        Background (\cbox{uddBlack}), 
        Vegetation (\cbox{uddGreen}), 
        Road (\cbox{uddGray}), 
        Vehicle (\cbox{uddBlue}), 
        Building (\cbox{uddRed}). 
        VDD:
Background (\cbox{vddBlack}),
Facade (\cbox{vddRed}),
Road (\cbox{vddYellow}),
Vegetation (\cbox{vddGreen}),
Vehicle (\cbox{vddBlue}),
Roof (\cbox{vddMagenta}),
Water (\cbox{vddCyan}).}
    \label{segresults2}
\end{figure}
\begin{table}[t]
    \centering
    \caption{Time and parameters comparison on OpenEarthMap.}
    \label{computational_cost}
    \resizebox{\linewidth}{!}{
    \begin{tabular}{l c c c c c c}
        \toprule
        Metric & \multicolumn{2}{c}{\cellcolor{red!10}{ClearCLIP}} & \multicolumn{2}{c}{\cellcolor{cyan!10}{ProxyCLIP}} & \multicolumn{2}{c}{\cellcolor{yellow!10}{SegEarth-OV}} \\
        & Static & SeeCo & Static & SeeCo & Static & SeeCo \\
        \midrule
        \midrule
        Parameters (M) & 149.62 & 149.87 & 235.43 & 235.67 & 149.91 & 150.16 \\
        Time (s/Image) & 0.023 & 0.518 & 0.050 & 0.961 & 0.128 & 1.911 \\
        \bottomrule
    \end{tabular}
    }
\end{table}

\begin{figure*}[t]
    \centering
    \includegraphics[width=\textwidth]{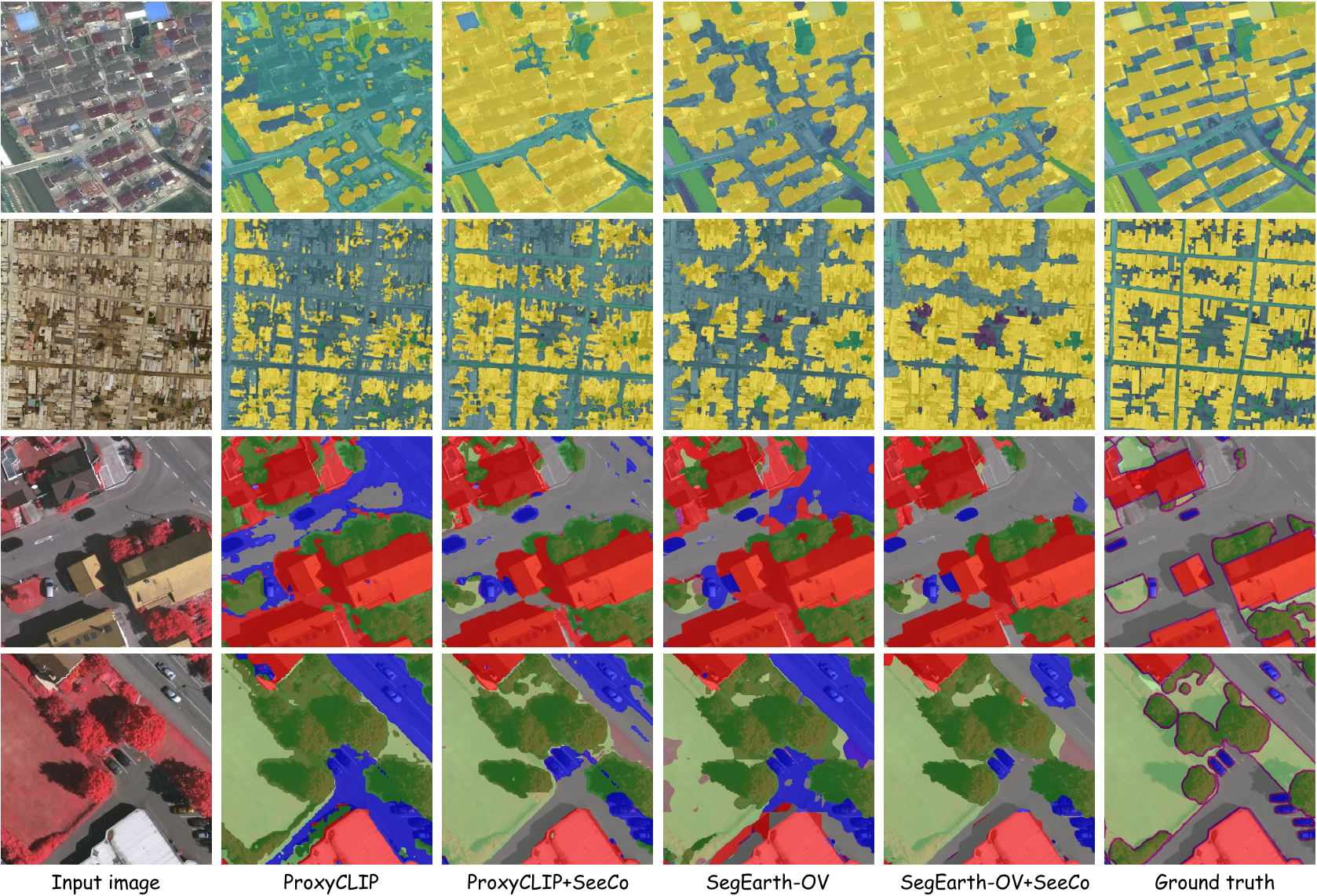}
    \caption{Qualitative results on OpenEarthMap and Vaihingen. 
        OpenEarthMap: 
        Background (\cbox{darkpurple}), 
        Grass (\cbox{emerald}), 
        Pavement (\cbox{darkbluegreen}), 
        Tree (\cbox{bluepurple}), 
        Water (\cbox{olivegreen}), 
        Building (\cbox{oliyellow}). 
        Vaihingen:
Impervious surface (\cbox{vaiGray}),
Building (\cbox{vaiRed}),
Low vegetation (\cbox{vaiLightGreen}),
Tree (\cbox{vaiForestGreen}),
Car (\cbox{vaiBlue}), Clutter (\cbox{vaiPurple}).
        }
    \label{segresults}
\end{figure*}
\begin{table*}[t]
    \centering
    \caption{Ablation study of SeeCo on eight remote sensing OVSS datasets. Static refers to the static inference mode used in existing methods. GCL and SCL indicate the proposed geometric and semantic consensus learning modules.}
    \label{ablationresults}
    \resizebox{\textwidth}{!}{
        \begin{tabular}{l cccc cccc cccc}
            \toprule
            \multirow{2}{*}{Dataset} 
            & \multicolumn{4}{c}{\cellcolor{red!10}{ClearCLIP}} 
            & \multicolumn{4}{c}{\cellcolor{cyan!10}{ProxyCLIP}} 
            & \multicolumn{4}{c}{\cellcolor{yellow!10}{SegEarth-OV}} \\
            & Static & +GCL & +SCL & +GCL+SCL & Static & +GCL & +SCL & +GCL+SCL & Static & +GCL & +SCL & +GCL+SCL \\
            \midrule 
            \midrule
            OpenEarthMap & 30.8 & 30.9 & {34.4} & {34.8} & 39.2 & 39.8 & {41.7} & {42.9} & 40.3 & {41.7} & 41.6 & {43.0} \\
            LoveDA & 31.0 & {33.5} & {35.9} & {36.6} & 36.4 & 36.6 & {39.9} & {40.3} & 36.9 & 36.8 & {38.7} & {39.4} \\
            iSAID & 18.2 & {19.0} & 18.6 & {19.5} & 21.4 & {22.4} & 21.7 & {22.7} & 21.7 & {23.0} & 21.1 & {22.2} \\
            Potsdam & 40.4 & {42.6} & 42.3 & {43.5} & 44.9 & {47.1} & {47.7} & {49.8} & 47.1 & {49.4} & 48.4 & {50.4} \\
            Vaihingen & 27.3 & 28.8 & {35.7} & {37.5} & 30.6 & 32.0 & {40.3} & {42.5} & 29.1 & 30.3 & {39.0} & {41.2} \\
            UAVid$^{\text{img}}$ & 36.7 & 37.4 & {40.1} & 40.0 & 40.9 & {44.3} & 41.8 & {44.3} & 42.5 & 42.7 & {45.3} & {44.9} \\
            UDD5 & 41.8 & 42.6 & {43.7} & {44.5} & 47.5 & {48.1} & {49.6} & {50.3} & 50.6 & {52.2} & 51.9 & {53.6} \\
            VDD & 38.8 & {40.4} & 38.4 & {40.9} & 43.9 & {46.3} & 43.4 & {46.7} & 45.3 & {45.6} & 44.3 & 45.3 \\
            \gr Avg & 33.1 & 34.4 & {36.1} & {37.2} & 38.1 & 39.6 & {40.7} & {42.4} & 39.2 & 40.2 & {41.3} & {42.5} \\
            \bottomrule
        \end{tabular}
    }
\end{table*}
As illustrated in Figure \ref{gain}, integrating SeeCo into ClearCLIP, ProxyCLIP, and SegEarth-OV can yield comprehensive performance improvements with nearly equivalent parameters. More specifically, inserting SeeCo into ClearCLIP and ProxyCLIP results in 4.1\% and 4.3\% gains in terms of mIoU, respectively, with comprehensive improvements across all datasets shown in Table \ref{table_overall}, confirming its effectiveness. Notably, incorporating SeeCo into ProxyCLIP achieves the state-of-the-art result, realizing a 1.7\% improvement over the previous methods, which further proves its powerful ability. Since general OVSS models rely on VLMs pre-trained on natural scenes, there exists a semantic domain gap with remote sensing images, resulting in unsatisfactory performance on certain satellite datasets, e.g., Vaihingen. In contrast, our SeeCo can alleviate this issue and achieve remarkable improvements of 10.2\% and 11.9\% on this challenging dataset, thereby demonstrating its superiority in narrowing the domain gap. Moreover, we visualize the feature space of existing methods and SeeCo using the t-SNE tool, as shown in Figure \ref{tsne}, which further demonstrates that SeeCo can eliminate land cover confusion.

Combining SeeCo with the remote sensing OVSS model still exhibits an overall performance improvement of 3.3\% in terms of mIoU, which proves that our method can still achieve stable enhancement even when integrated into the SOTA model, further demonstrating its superiority. Furthermore, we perform a qualitative visualization comparison for different methods, as illustrated in Figures \ref{segresults2} and \ref{segresults}. SeeCo effectively alleviates incomplete segmentation and missing detection issues in land-cover regions. Next, we compare the running cost in Table \ref{computational_cost}, which exhibits a slight increase in time cost and parameters, proving its efficiency.

\begin{table}[t]
    \centering
    \caption{Ablation study of GCL and SCL. Static: static inference, MIM: masked image modeling, PL: pseudo labels, Text: text prompts, Vision: vision prompts.} 
    \label{table_ablation_combined}
    \resizebox{\linewidth}{!}{
        \begin{tabular}{l cccc cccc}
            \toprule
            \multirow{2}{*}{Dataset} 
            & \multicolumn{4}{c}{\cellcolor{red!10}{GCL Ablation}} 
            & \multicolumn{4}{c}{\cellcolor{cyan!10}{SCL Ablation}} \\
            \cmidrule(lr){2-5} \cmidrule(lr){6-9}
            & Static & MIM & PL & GCL & Static & Text & Vision\&Text & SCL \\
            \midrule 
            \midrule 
            OpenEarthMap & 39.2 & 39.8 & 39.4 & \textbf{39.8} & 39.2 & 41.0 & 41.1 & \textbf{41.7} \\
            LoveDA & 36.4 & 36.5 & 36.5 & \textbf{36.6} & 36.4 & 39.2 & \textbf{40.2} & 39.9 \\
            iSAID & 21.4 & 21.3 & 22.4 & \textbf{22.4} & 21.4 & \textbf{22.0} & 21.8 & 21.7 \\
            Potsdam & 44.9 & 45.7 & \textbf{47.5} & 47.1 & 44.9 & 45.5 & 46.2 & \textbf{47.7} \\
            Vaihingen & 30.6 & 30.7 & 31.9 & \textbf{32.0} & 30.6 & 35.9 & 39.4 & \textbf{40.3} \\
            UAVid$^{\text{img}}$ & 40.9 & 41.6 & 42.2 & \textbf{44.3} & 40.9 & 41.7 & \textbf{43.4} & 41.8 \\
            UDD5 & 47.5 & 47.7 & 47.9 & \textbf{48.1} & 47.5 & 48.3 & 49.4 & \textbf{49.6} \\
            VDD & 43.9 & 44.9 & 44.8 & \textbf{46.3} & \textbf{43.9} & 43.6 & 41.9 & 43.4 \\
            \gr Avg & 38.1 & 38.5 & 39.1 & \textbf{39.6} & 38.1 & 39.7 & 40.4 & \textbf{40.8} \\
            \bottomrule
        \end{tabular}
    }
\end{table}

\begin{figure}[t]
    \centering

    \includegraphics[width=.95\linewidth]{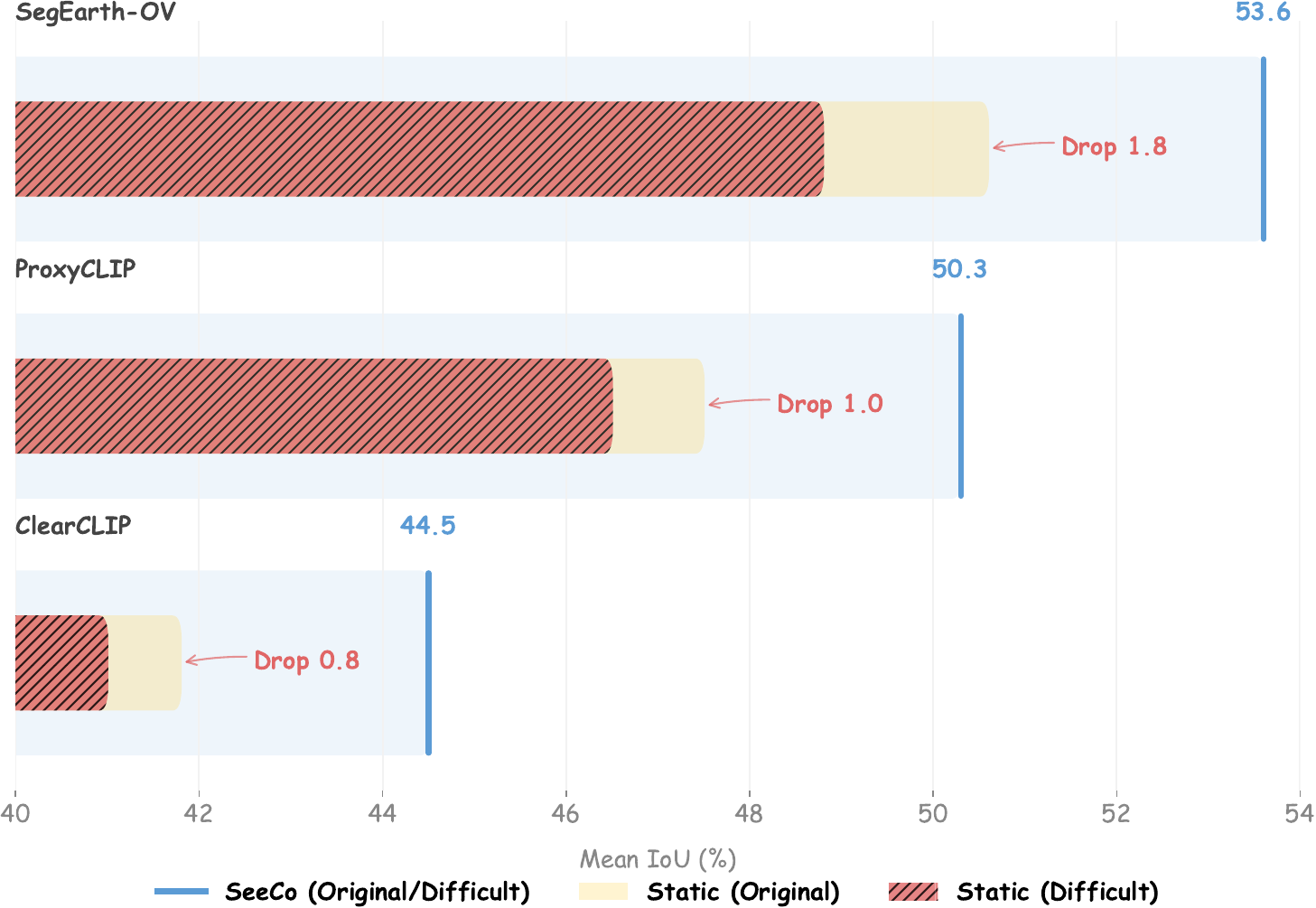}
    \caption{Performance degradation comparison of different open-vocabulary semantic segmentation methods in challenging scenarios with multi-view observations.}
    \label{drop}
\end{figure}

\subsection{Ablation Study}

\paragraph{Effect of each component:} We conduct detailed ablation studies on three basic models, as shown in Table \ref{ablationresults}. For ProxyCLIP, its segmentation performance of static inference is 38.1\%. On the basis of this, introducing the geometric and semantic consensus learning mechanisms results in performance gains of 1.5\% and 2.6\%, respectively, demonstrating that different observations and more diverse textual descriptions can effectively improve the cross-modal segmentation capabilities of the model. We further compare the ProxyCLIP with SeeCo on UDD5 under various observation views (0, $\frac{\pi}{2}$, ${\pi}$, $\frac{3\pi}{2}$). As shown in Figure \ref{drop}, the results indicate that variation in observation views will lead to a performance degradation, validating the necessity of GCL.

When both geometric and semantic consensus learning mechanisms are combined, the model realizes its best performance of 42.4\% with a total improvement of 4.3\% in terms of mIoU, demonstrating the complementary advantages of the two mechanisms. Similarly, building upon SegEarth-OV, which is designed for RSIs via spatial enhancement, introducing geometric consensus still yielded a 1\% performance improvement, proving its effectiveness. Using the semantic consensus learning mechanism can result in a more significant gain of 2.1\%. Since SegEarth-OV does not perform additional processing on the text encoder, SCL significantly improved performance by mitigating this drawback. The above experimental results demonstrate the effectiveness of each component in SeeCo.

\paragraph{Different self-supervised learning strategies in GCL:} We make a comparison of different self-supervised learning strategies. MIM and PL indicate masked image modeling strategy \cite{MaskSSL} and pseudo labels strategy, respectively. The former randomly masks the image and uses the reconstruction error for supervisory signals. The latter performs a maximum value operation on the predicted images under $K$ views to generate pseudo-labels, which are used to optimize the original predictions. As shown in Table \ref{table_ablation_combined}, applying MIM can also obtain a 1\% gain of mIoU by guiding the model to adapt to the current scene. However, it neglects the observation property of remote sensing images, resulting in suboptimal results. 
In contrast, GCL suppresses false alarms and improving the integrity of foreground activation.

\paragraph{Different prompting strategies in SCL:} We further explore different prompt strategies in the proposed SCL strategy in Table \ref{table_ablation_combined}. Using text prompts to acquire rich descriptions achieves remarkable segmentation results with a 39.7\% mIoU, which outperforms the static model by 1.6\%, confirming the effectiveness of synonyms. Integrating text and vision prompts with the adaptive context recalibration module, it achieves satisfactory results with 40.4\% mIoU.

\begin{figure}[t]
    \centering

    \includegraphics[width=.95\linewidth]{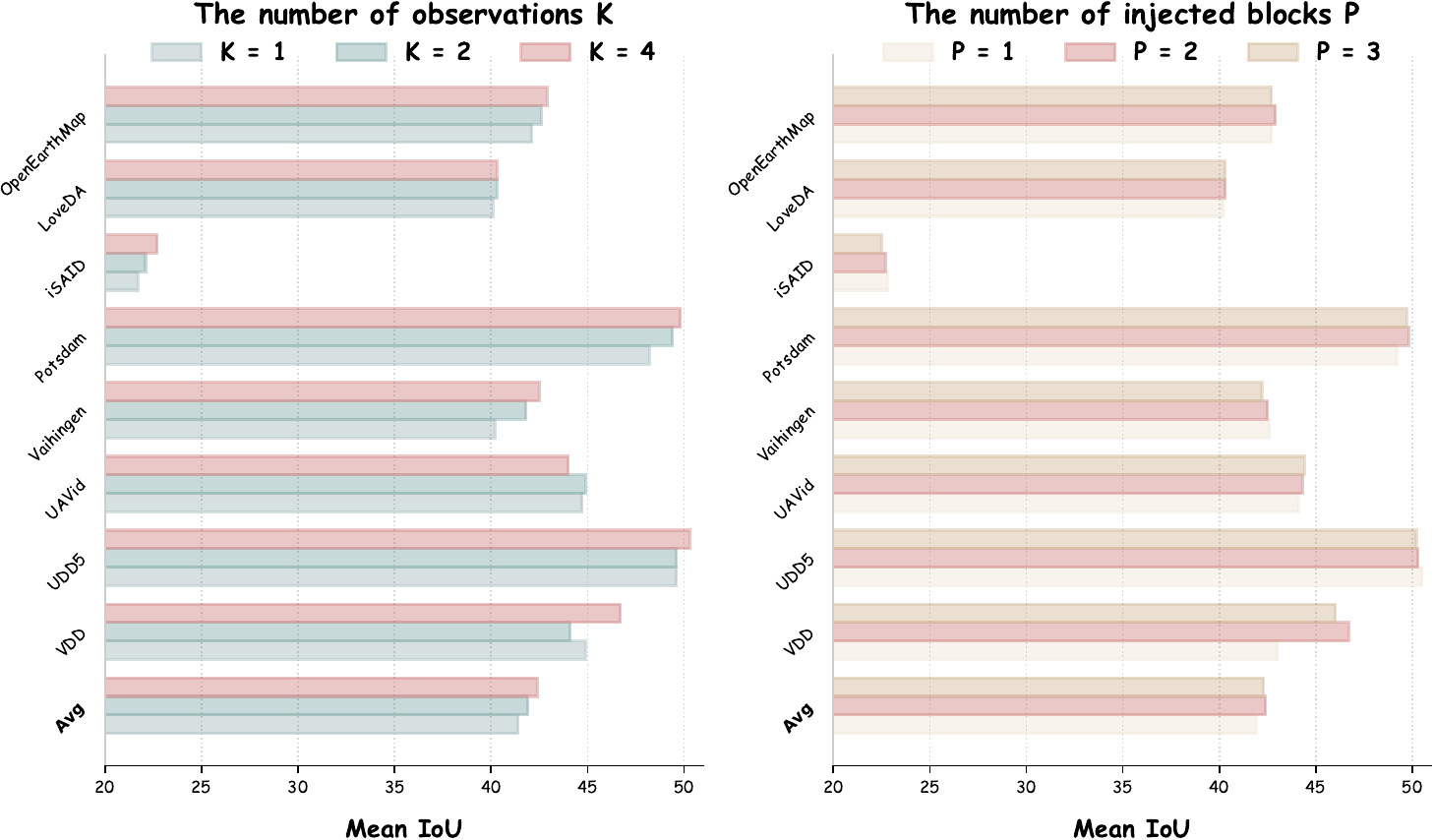}
    \caption{Hyperparameter analysis of the number of multi-view observations $K$, and the number of injected blocks $P$ on eight remote sensing OVSS datasets.}
    \label{param}
\end{figure}

\paragraph{Hyperparameter analysis:} We study a comprehensive analysis of the hyperparameter in SeeCo, especially for the number of multi-view observations $K$ and the number of injected blocks $P$. As shown in Figure \ref{param}, our SeeCo is robust to the hyperparameters with slight variations in terms of segmentation results. It can be seen that adopting more observations results in a performance gain due to capturing more spatial cues, where $K$=4 achieves the best performance. Injecting the consensus into the last 2 blocks realizes a better result. More injected blocks will lead to an extra computational cost, yet without a performance gain.

%% file: sec/5_Conclusion.tex
\section{Conclusion}
In this paper, we proposed a plug-and-play framework for remote sensing open-vocabulary semantic segmentation (OVSS) called Seeking Consensus (SeeCo), which takes a detour from existing static inference paradigms and recalibrates the OVSS model online for each scene. Specifically, SeeCo recalibrated arbitrary existing OVSS models on-the-fly by jointly learning geometric and semantic consensus targets. The former was acquired from multi-view observations with different rotated views, and the latter was built through rich textual description from LLMs. Both consensus were injected using an online consensus injector (OCI) to realize scene-adaptive parameter calibration, thereby enhancing segmentation performance. Extensive experiments on eight remote sensing OVSS benchmarks demonstrate its superiority.